\documentclass[letterpaper, 10 pt, journal, twoside]{IEEEtran}

\IEEEoverridecommandlockouts                              
\usepackage{times}
\usepackage{amsmath, amssymb, amsfonts}
\usepackage{bm}
\usepackage{textcomp}
\usepackage{graphicx}
\usepackage{epsfig}
\usepackage[export]{adjustbox}
\usepackage{booktabs}
\usepackage{float}
\usepackage[dvipsnames,table]{xcolor}  
\usepackage{colortbl}           
\usepackage{multirow}
\usepackage{array}
\usepackage{tabularx}
\usepackage{stfloats}
\usepackage[caption=false,font=normalsize,labelfont=sf,textfont=sf]{subfig}
\usepackage[ruled]{algorithm}
\usepackage{algpseudocode}
\usepackage{url}
\usepackage[bookmarks=true]{hyperref}
\usepackage{pgf}
\usepackage{import}
\usepackage{soul}
\usepackage{verbatim}
\usepackage{lettrine}
\usepackage{url}
\usepackage{ctable}
\usepackage[bookmarks=true]{hyperref}
\usepackage{makecell}

\usepackage[sort&compress,numbers]{natbib}
\usepackage{balance}
\def\BibTeX{{\rm B\kern-.05em{\sc i\kern-.025em b}\kern-.08em
  T\kern-.1667em\lower.7ex\hbox{E}\kern-.125emX}}
\usepackage{packages/sparo_acronyms}
\usepackage{packages/sparo_math}
\usepackage{packages/sparo_SIunits}
\usepackage{packages/sparo_misc}

\usepackage[symbol]{footmisc}

\definecolor{gl}{HTML}{008000}

\begin{document}
\title{
MSG-Loc: Multi-Label Likelihood-based\\Semantic Graph Matching for Object-Level \\Global Localization
    
}


\author{Gihyeon Lee$^{1}$, Jungwoo Lee$^{1}$, Juwon Kim$^{1}$, Young-Sik Shin$^{2\dagger}$, and Younggun Cho$^{1\dagger}$
\thanks{\hspace*{-1.2em}Manuscript received: June 30, 2025; Accepted: November 18, 2025. This letter was recommended for publication by Editor Sven Behnke upon evaluation of the Associate Editor and Reviewers' comments. This work was supported by Institute of Information \& communications Technology Planning \& Evaluation (IITP) grant funded by the Korea government (MSIT) (RS-2022-II220448), National Research Foundation of Korea (NRF) grant (RS-2025-02217000 and RS-2025-24803365), Industrial Strategic Technology Development Program (no.20018745) funded by the Ministry of Trade, Industry \& Energy (MOTIE, Korea), and National Research Council of Science \& Technology as part of the project titled ``Development of Core Technologies for Robot General Purpose Task Artificial Intelligence Framework" (NK254G). \textit{(Corresponding Authors: Young-Sik Shin and Younggun Cho)}} 
\thanks{\hspace*{-1.2em}$^{1}$Gihyeon Lee, Jungwoo Lee, Juwon Kim, and Younggun Cho are with Electrical and Computer Engineering, and INHA Future Mobility IPCC, Inha University, Incheon 22212, South Korea (e-mail: leekh951@inha.edu; pihsdneirf@inha.edu; marimo117@inha.edu, yg.cho@inha.ac.kr).}
\thanks{\hspace*{-1.2em}$^{2\dagger}$Young-Sik Shin is with the School of Mechanical Engineering, Kyungpook National University, Daegu, South Korea (e-mail: yshin86@knu.ac.kr).}
\thanks{\hspace*{-1.2em}Digital Object Identifier (DOI): see top of this page.}
}

\maketitle
\begin{abstract}
Robots are often required to localize in environments with unknown object classes and semantic ambiguity. 
However, when performing global localization using semantic objects, high semantic ambiguity intensifies object misclassification and increases the likelihood of incorrect associations, which in turn can cause significant errors in the estimated pose.
Thus, in this letter, we propose a multi-label likelihood-based semantic graph matching framework for object-level global localization.
The key idea is to exploit multi-label graph representations, rather than single-label alternatives, to capture and leverage the inherent semantic context of object observations.
Based on these representations, our approach enhances semantic correspondence across graphs by combining the likelihood of each node with the maximum likelihood of its neighbors via context-aware likelihood propagation.
For rigorous validation, data association and pose estimation performance are evaluated under both closed-set and open-set detection configurations.
In addition, we demonstrate the scalability of our approach to large-vocabulary object categories in both real-world indoor scenes and synthetic environments.
Project Page: https://sparolab.github.io/research/msg-loc/.
\end{abstract}

\section{Introduction}
\IEEEPARstart{G}{lobal} localization on a prior map, which estimates the pose of a robot from current sensor observations without any prior pose information, is an essential task in robotics~\cite{Toward}. 
This capability is important for \ac{SLAM} tasks such as relocalization after tracking failures and loop closures~\cite{orbslam2, oaslam}.
While recent feature-based methods~\cite{sun2021loftr} improve robustness to viewpoint and appearance changes, their performance still degrades under severe variations. In contrast, object-level approaches maintain more reliable global localization in such conditions.

\begin{figure}[t]
	\centering
	\def\width{0.49\textwidth}%
    {%
        \includegraphics[clip, trim= 0 0 0 0, width=\width]{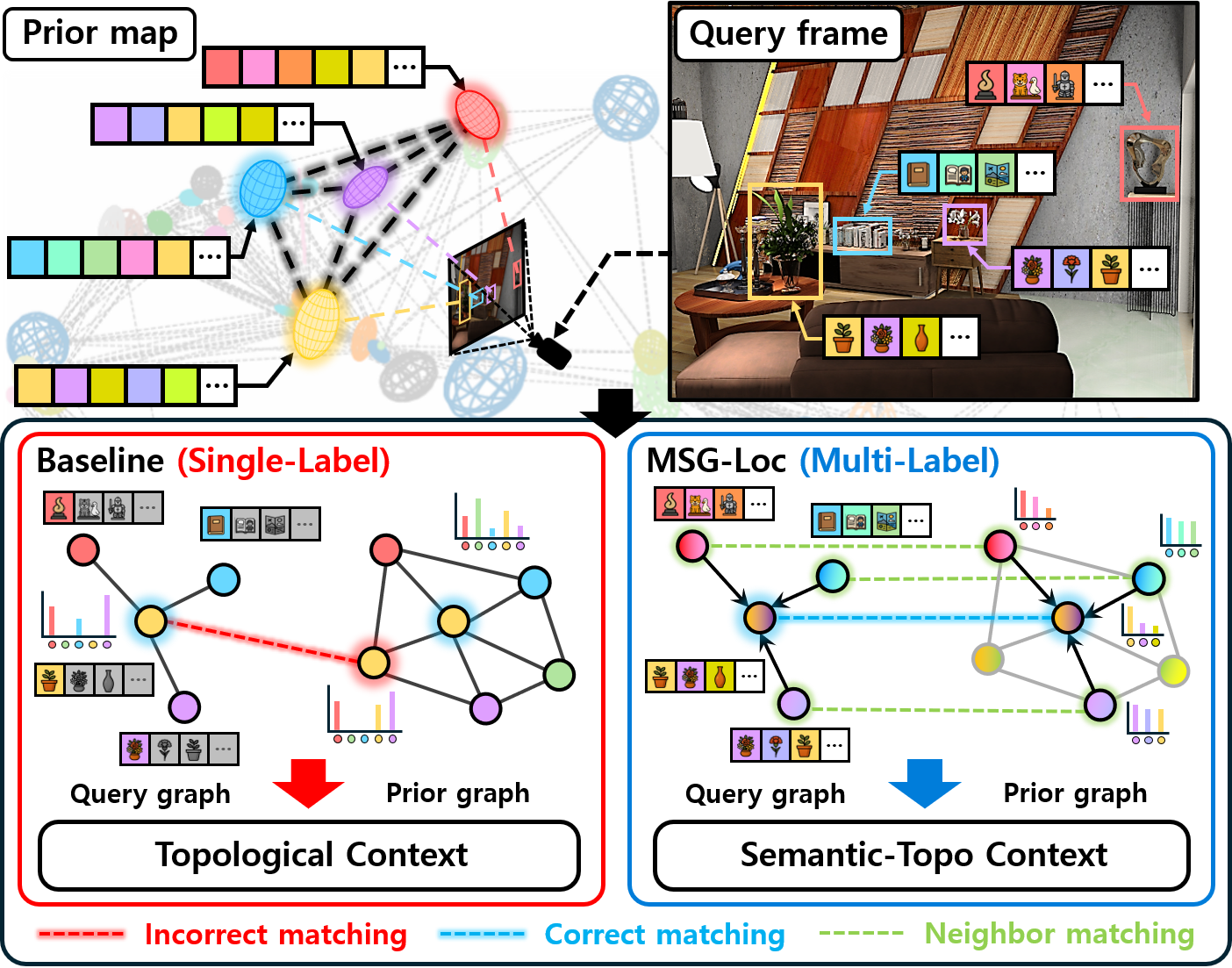}
    }
    \vspace{-0.6cm}
    \caption{\textbf{MSG-Loc.} Our method preserves multi-label hypotheses for both landmarks and observed objects, and estimates their likelihoods.
    It propagates context-aware likelihoods and calculates similarity across the semantic graph to mitigate viewpoint-dependent misclassifications.
    This enables reliable object-level global localization even in scenes of severe semantic ambiguity.
}	
\vspace{-0.5cm}
\label{fig:fig1}
\end{figure}
However, most existing object-based methods rely on single-label predictions, neglecting semantic uncertainty and label inconsistency prevalent in practical scenarios~\cite{viewpoint}.
In particular, the single-label assumption has limitations in open-set environments where unknown objects are present.
In this context, prior studies~\cite{objectdhd, semantictopoloop, eaoslam++, xview, semantichistogram, goreloc} have introduced semantic graph-based matching methods leveraging relationships between adjacent objects, but struggle to capture semantic variations of individual objects induced by partial observations or viewpoint changes. 
For instance, the graph kernel-based approach~\cite{goreloc} constructs kernel descriptors that account for object detection uncertainty in the prior graph.
Nevertheless, since it assigns only a single label per object in query frames, it is limited in handling the semantic ambiguity inherent in open-set observations.
Additionally, fixed descriptors derived from sparse object distributions are sensitive to semantic inconsistencies, often leading to matching errors.

Recently, \ac{VLM}-based methods \cite{cliploc, clipclique} have been explored to recognize out-of-distribution classes. 
However, these approaches have limited applicability in real-world scenarios as they assign to each landmark a single text label, which is typically generated manually or via offline captioning.
Therefore, it is necessary to develop flexible subgraph matching approaches capable of integrating node-level uncertainty and adapting to sparse and ambiguous observations in open-set environments.

In this letter, we propose \textit{MSG-Loc}, a novel semantic graph matching framework based on multi-label likelihoods for global localization, as shown in Fig.~\ref{fig:fig1}.
The key idea is to preserve multi-label hypotheses observed from multiple viewpoints during map construction and to exploit these hypotheses during querying, thereby mitigating the semantic ambiguity inherent to single-label predictions.
To this end, we define a multi-label likelihood for each node-pair and leverage maximum-likelihood propagation from neighboring matches, which leads to more reliable candidate selection for global localization.
The main contributions of this work are summarized as follows:
\begin{itemize}
    \item \textbf{Semantic Disambiguation:} \
    To the best of our knowledge, this is the first approach to capture semantic uncertainty and consistency by exploiting multi-label likelihoods to estimate inter-object correspondences.
    \item \textbf{Context-Aware Stability:} \
    The proposed method propagates multi-label likelihoods via 1-hop neighbors in the semantic graph, enabling stable data association and pose estimation under sparse and ambiguous observations.
    \item \textbf{Scalability to Large Class Sets:} \
    Our method achieves reliable performance over a wide vocabulary of object classes, as the multi-label likelihood is not constrained by the size or composition of the label set.
    \item \textbf{Cross-Paradigm Compatibility:} \
    The proposed method shows effectiveness across multiple paradigms, including open-set and closed-set detection as well as zero-shot and supervised classification tasks.
\end{itemize}

\section{Related Work}
\subsection{Object-based SLAM}
Unlike traditional feature-based methods~\cite{orbslam2}, object-based approaches leverage high-level semantics to enhance localization accuracy and robustness. 
In semantic \ac{SLAM} frameworks, accurate object representation and recognition form the basis for building semantically consistent maps.

SLAM++~\cite{slam++} introduced an object-oriented \ac{SLAM} approach that utilizes known object models. 
However, this approach lacks generalization to unknown objects. 
To address this limitation, CubeSLAM~\cite{cubeslam} proposed monocular object \ac{SLAM} by estimating 3D cuboids directly from 2D bounding boxes. 
\citet{mono} further enhanced this method by integrating planar surfaces with cuboid objects, resulting in improved localization accuracy and richer environmental representation.

In recent years, dual quadrics have emerged as a general and compact representation of objects within \ac{SLAM} frameworks. 
Since QuadricSLAM~\cite{quadricslam} formalized dual quadrics as compact 3D landmarks, subsequent works have adopted this representation for object-level \ac{SLAM} owing to its analytic projection model and efficient parameterization.
For example, EAO-SLAM~\cite{eaoslam} proposed a semi-dense object-based framework that leverages an ensemble strategy with statistical tests to improve data association.
QISO-SLAM~\cite{qisoslam} introduced a more precise observation model by incorporating instance segmentation-derived contour points. Recently, VOOM~\cite{voom} proposed a hierarchical landmark representation combining high-level objects and low-level points, demonstrating superior localization performance over traditional object-based and feature-based methods.
\subsection{Semantic Graph Matching for Global Localization}
Despite these advancements, previous approaches focusing on dual quadric-based mapping and localization have not adequately addressed crucial aspects such as relocalization and loop detection, especially after tracking loss.
To address these limitations, OA-SLAM~\cite{oaslam} fused sparse points with ellipsoidal object models, enabling camera relocalization through object-wise matching from viewpoints where the point-only baseline~\cite{orbslam2} fails.

Beyond this relocalization method, graph-based approaches incorporating topological information have gained prominence for reliable localization in object-level maps. 
\citet{semantichistogram} introduced \ac{SH}, a histogram-of-label-paths descriptor computed on semantic graphs that captures higher-order context and enables faster graph matching than the random-walk approach~\cite{xview}.
Additionally, \citet{objectdhd} developed directional histogram descriptors tailored for object-based loop closure, demonstrating reliable performance even with extensive viewpoint differences.
\citet{eaoslam++} expanded the object map proposed in EAO-SLAM~\cite{eaoslam} into a topological map framework, achieving multi-scene matching and semantic descriptor-driven relocalization.
SemanticTopoLoop~\cite{semantictopoloop} proposed multi-level verification strategies, combining object data association and quadric-level topological maps for semantic loop closure, effectively reducing false positives. 

However, these graph-based methods have mainly relied on topological context, overlooking the semantic ambiguity inherent to each object.
To address this limitation, GOReloc~\cite{goreloc} integrated the graph kernel-based matching strategy of \citet{stumm_graphkernel} with label statistics accumulated across keyframes.
Nevertheless, the aforementioned methods still depend on single-label predictions per object in query frames, failing to fully capture semantic uncertainty at inference time.

In contrast to previous methods, we propose a multi-label likelihood-based graph matching framework that incorporates both the semantic context of candidate labels and the topological context of the graph to establish object correspondences.
To estimate the multi-label likelihood, we employ the multi-label detection frequencies from the map and the normalized confidence distributions from the input frame.
\section{MSG-Loc}
\subsection{Problem Definition}
\begin{figure*}[t]
	\centering
	\def\width{0.95\textwidth}%
    {%
        \includegraphics[clip, trim= 0 0 0 0, width=\width]{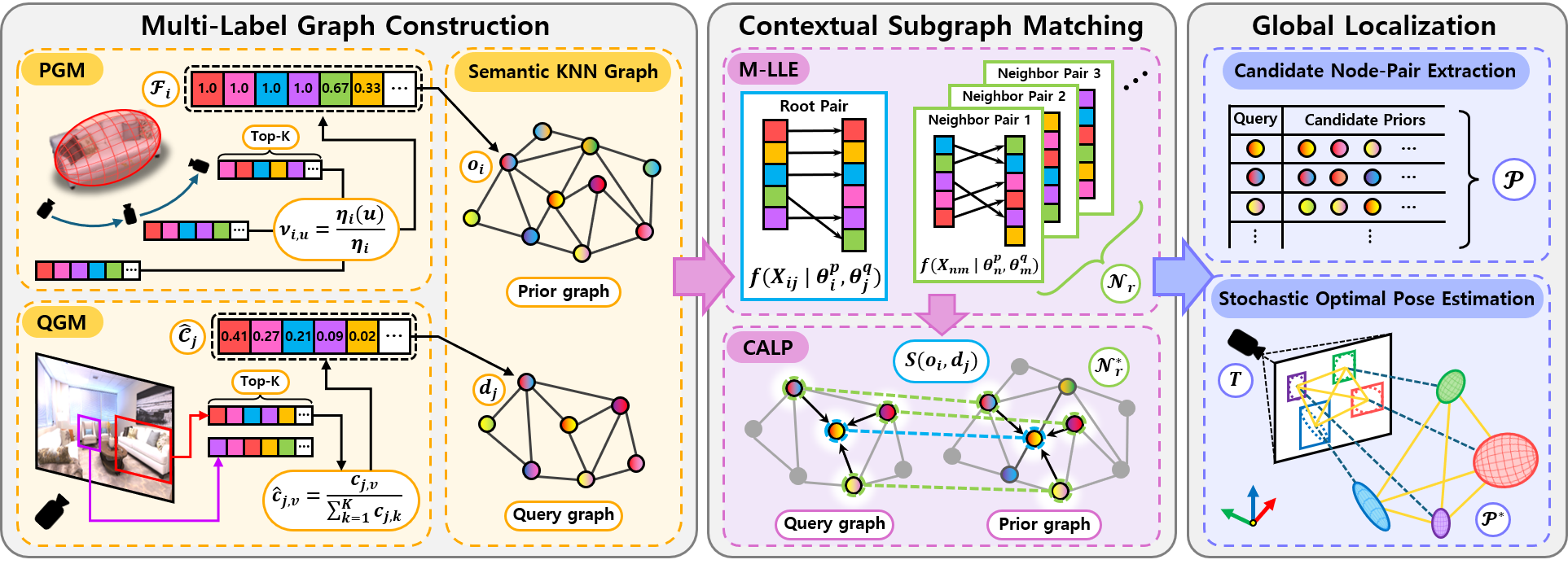}
    }
    \vspace{-0.3cm}
     \caption{\textbf{Overview of our method.}
In the Multi-Label Graph Construction phase, we generate a k-nearest neighbor (KNN)-based semantic graph, embedding multi-label detection frequencies and normalized confidence scores via the Prior Graph Management (PGM) and Query Graph Management (QGM) modules.
Multi-Label Likelihood Estimation (M-LLE) then computes semantic likelihoods from these attributes for both landmarks and observations.
Context-Aware Likelihood Propagation (CALP) propagates the maximum likelihood from neighboring nodes to the root node, calculating its similarity score to ensure contextual consistency.
Finally, candidate node-pairs are extracted, and the stochastic optimal pose estimation module determines the camera pose.
             } 
    \label{fig:fig2}
    \vspace{-0.5cm}
\end{figure*}

Our objective is to match a query graph $\mathcal{G}^q$ to its corresponding subgraph in a prior graph $\mathcal{G}^p$ for global localization, as illustrated in Fig.~\ref{fig:fig2}, where the superscripts $q$ and $p$ denote the query and prior domains, respectively.

For this purpose, the \ac{PGM} and \ac{QGM} modules construct prior nodes $\mathcal{O}^p$ and query nodes $\mathcal{D}^q$ from synchronized RGB-D input, each characterized by multi-label attributes and defined in 3D Euclidean space.
Each $i^{\text{th}}$ prior object node $o_i\in\mathcal{O}^p$ contains dual quadric parameters $\mathbf{q}^p_i = [\mathbf{p}^p_i, \mathbf{r}_i, \mathbf{s}_i]$ and multi-label detection frequencies $\mathcal{F}_i$, where $\mathbf{p}^p_i$, $\mathbf{r}_i$, and $\mathbf{s}_i$ denote the position, rotation, and scale vectors, respectively.
For each input frame, the $j^{\text{th}}$ query node $d_j \in \mathcal{D}^q$ comprises a bounding box $B_j$ from the object detector, a normalized multi-label confidence $\hat{\mathcal{C}}_j$ derived from the classifier output ${\mathcal{C}}_j$, and a 3D position $\mathbf{p}^q_j$ estimated using $B_j$ and the depth image.

Using these graph formulations, candidate node-pairs $\mathcal{P}$ are extracted from $\mathcal{G}^p$ and $\mathcal{G}^q$ by integrating topological structure and multi-label semantic features through the \ac{M-LLE} and \ac{CALP} modules.
Finally, the camera pose $T$ is estimated by iteratively determining the optimal node correspondences $\mathcal{P}^{*}$.
\subsection{Graph Construction via Multi-Label Prediction}\label{sec:3-B}
By utilizing multi-label rather than single-label statistics, the prior graph forms the basis for stable and consistent object descriptions, even in the presence of transient detection errors and viewpoint variations. Likewise, by adopting multi-label confidence distributions, the query graph captures latent semantic labels beyond what single-label approaches represent at the object level.
\subsubsection{Prior Graph Management}
Across multiple keyframes, the prior graph $\mathcal{G}^p$ preserves semantic consistency of individual objects by assigning accumulated multi-label detection counts to the node set $\mathcal{O}^p$.
Each prior object node $o_i = \{\mathbf{q}^p_i, \mathcal{F}_i\}$ is represented by $\mathbf{q}^p_i$, the optimized 3D object.
Specifically, the set $\mathcal{F}_i = \{(l^p_{i,u}, \nu_{i,u})\}_{u=1}^U$ is constructed by aggregating top-$K$ predicted labels across keyframes, where $U$ denotes the number of unique labels of $o_i$.
The per-label detection frequency $\nu_{i,u}$ of the $u^{\text{th}}$ unique label $l^p_{i,u}$ is defined relative to all detections of $o_i$, as follows:
\begin{equation}
\nu_{i,u} = \frac{\eta_{i}(u)}{\eta_i},
\label{eq:1}
\end{equation}
where $\eta_{i}$ is the total number of detections of $o_i$, and $\eta_{i}(u)$ counts detections where $l^p_{i,u}$ appears among the predicted labels.
After node generation, the prior graph $\mathcal{G}^p = (\mathcal{O}^p, \mathcal{E}^p)$ is constructed by merging, for each keyframe, all \ac{KNN} edges $\mathcal{E}^p$ formed based on $\mathbf{p}^p_i$.
\subsubsection{Query Graph Management}
For each input frame, the query graph encapsulates semantic uncertainty by assigning normalized top-$K$ multi-label confidence distributions to each detected object node in the set $\mathcal{D}^q$.
Each query detection node $d_j = \{B_j, \mathbf{p}^q_j, \hat{\mathcal{C}}_j\}$ is represented as a bounding box $B_j$ with the position vector $\mathbf{p}^q_j$. 
Notably, the set $\hat{\mathcal{C}}_j = \{(l^q_{j,v}, \hat{c}_{j,v})\}_{v=1}^K$ contains the $v^{\text{th}}$ top label $l^q_{j,v}$ and its corresponding normalized confidence score $\hat{c}_{j,v}$, where $K$ is the number of top-ranked labels retained for each detection.
The normalization process is defined as follows:
\begin{equation}
\hat{c}_{j,v} = \frac{c_{j,v}}{\sum_{k=1}^{K} c_{j,k}},
\label{eq:2}
\end{equation}
where $c_{j,v}$ denotes the raw confidence score for $l^q_{j,v}$, obtained from $(l^q_{j,v}, c_{j,v}) \in \mathcal{C}_j$.
After generating the nodes, the query graph $\mathcal{G}^q = (\mathcal{D}^q, \mathcal{E}^q)$ is constructed by forming \ac{KNN} edges $\mathcal{E}^q$ based on $\mathbf{p}^q_j$.
\subsection{Multi-Label Likelihood Estimation}\label{sec:3-C}
\begin{figure}[t]
	\centering
	\def\width{0.465\textwidth}%
    {%
        \includegraphics[clip, trim= 30 0 10 0, width=\width]{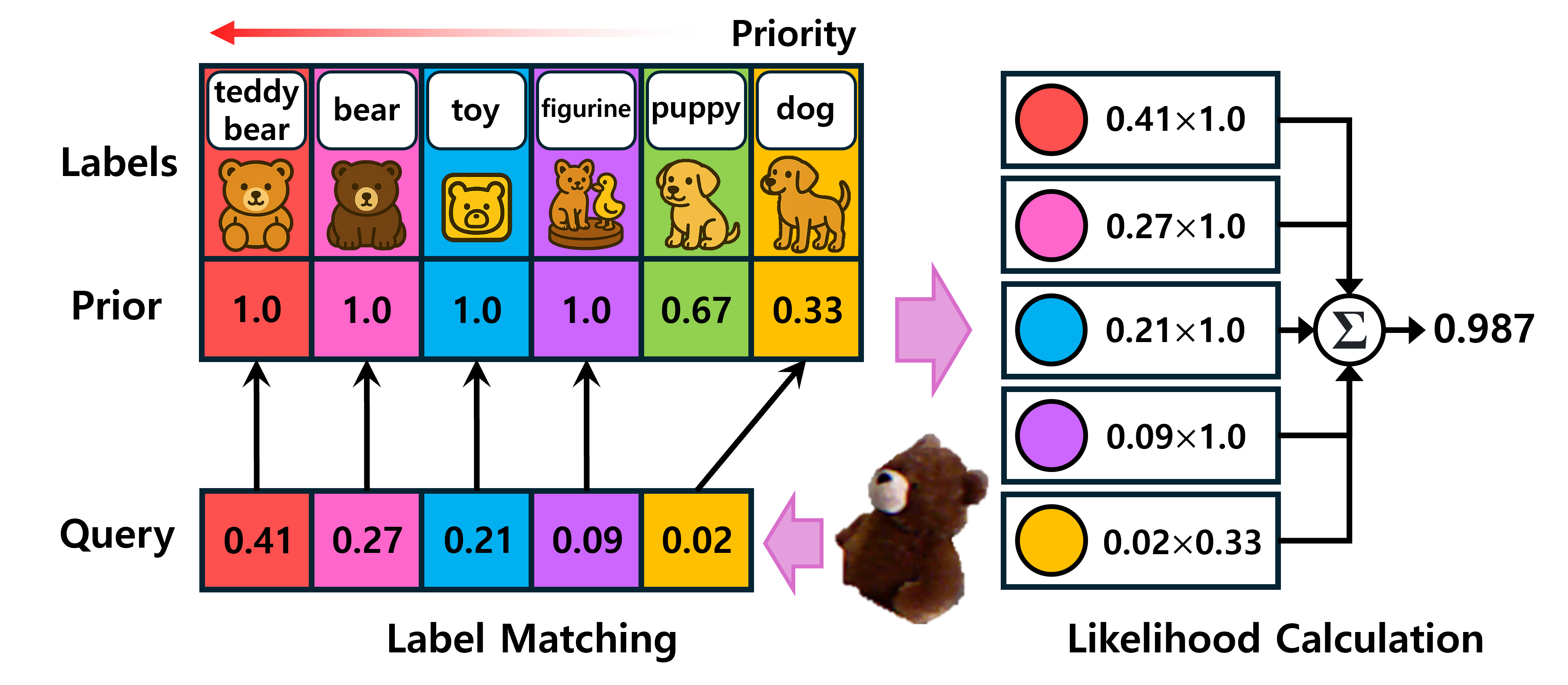}
    }
        \vspace{-0.3cm}
        \caption{\textbf{Multi-label likelihood estimation algorithm.}
This figure illustrates how M-LLE computes semantic likelihoods between landmark and observation nodes by aggregating detection frequencies and normalized confidence scores.
The probabilistic estimation of likelihoods enables data association that is agnostic to the object classification paradigm within our framework.
}
\label{fig:fig3}
\vspace{-0.5cm}
\end{figure}

Ambiguous detections often assign non-negligible confidence scores to several conceptually similar labels for a single object, such as ‘cup’, ‘mug’, and ‘glass’.
Constraining multi-label candidates to a single deterministic label overlooks this semantic uncertainty and often results in mismatches between visually similar objects.
So, we focus on leveraging the semantic context among candidate labels as a key feature for object matching, as illustrated in Fig.~\ref{fig:fig3}.

Under the assumption of pairwise independence for tractability, we formulate the multi-label likelihood in terms of empirical evidence $(l_{i,u}^p, \nu_{i,u})$ and perceptual cues $(l_{j,v}^q, \hat{c}_{j,v})$ as follows:
\begin{equation}
f\bigl(X_{ij} \mid \theta_i^p, \theta_j^q\bigr) \equiv \sum_{u=1}^{U}\sum_{v=1}^{K} \mathbb{I}\bigl(l_{i,u}^p = l_{j,v}^q\bigr) \cdot \nu_{i,u} \cdot \hat{c}_{j,v},
\label{eq:3}
\end{equation}
where $\theta^p_i$, $\theta^q_j$, and $X_{ij}$ denote parameters associated with $\mathcal{F}_i$, $\hat{\mathcal{C}}_j$, and a random variable representing the correspondence between $o_i$ and $d_j$, respectively.
The indicator function $\mathbb{I}(\cdot)$ returns 1 if the given condition is true, and 0 otherwise. 
That is, the indicator function restricts the summation to only those label pairs that are identical, so that the likelihood reflects the degree of alignment between the empirical and perceptual label distributions.

Importantly, since \ac{M-LLE} does not rely on a fixed set of class categories, it is not constrained by the size or composition of the label vocabulary.
As a result, it is designed to be detector-agnostic and remains applicable to both open-set and closed-set configurations.
\subsection{Context-Aware Likelihood Propagation}\label{sec:3-D}
While the \ac{M-LLE} evaluates node-pairs independently for efficiency, it may overlook contextual correlations across neighboring objects. Thus, it is necessary to enable correct pair selection by jointly considering semantic alignment and topological consistency across the entire graph via \ac{CALP}.

First, we define the set of neighboring node-pairs $\mathcal{N}_r = \{(o_{n}, d_{m})\}$ for each root node-pair $(o_{i}, d_{j})$, where $o_{n}$ and $d_{m}$ denote the $n^{\text{th}}$ and $m^{\text{th}}$ 1-hop neighbors of $o_{i}$ and $d_{j}$ in the prior and query graphs, respectively.
To reflect the similarity of relative distances to the root node in both graphs, the corresponding weight $w_{nm}$ is defined as:
\begin{equation}
w_{nm} = \frac{1}{1 + |\delta^p_{n} - \delta^q_{m}|},
\label{eq:4}
\end{equation}
where $\delta^p_{n} = \|\mathbf{p}^p_i-\mathbf{p}^p_n\|_2$ and $\delta^q_{m} = \|\mathbf{p}^q_j-\mathbf{p}^q_m\|_2$ are the Euclidean distances from the root node to each neighbor node in the prior and query graphs, respectively. The vectors $\mathbf{p}^p_n$ and $\mathbf{p}^q_m$ represent the positions of nodes $o_n$ and $d_m$.

Then, the best neighboring node-pair set $\mathcal{N}^*_r$ is selected from $\mathcal{N}_r$ by maximizing the weighted neighbor likelihood for each neighboring query node as follows:
\begin{equation}
\mathcal{N}^*_r = \bigcup_{m=1}^M \ \operatorname*{arg\,max}_{(o_{n}, d_{m})\in\mathcal{N}_r} w_{nm} f\bigl(X_{nm} \mid \theta_{n}^p, \theta_{m}^q\bigr),
\label{eq:5}
\end{equation}
where $M$ denotes the number of neighboring nodes connected to the root node in the query graph.
The weighted likelihoods of the neighbors identified by $\mathcal{N}^*_r$ are propagated to the root node-pair, and the similarity score is calculated as follows:
\begin{equation}
\begin{aligned}
S(o_i,d_j) =\ &f\bigl(X_{ij} \mid \theta_i^p, \theta_j^q\bigr) \\ & +\frac{1}{|\mathcal{N}^*_r|} \sum_{(o_{n}, d_{m})^*\in \mathcal{N}^*_r} w_{nm} f\bigl(X_{nm} \mid \theta_{n}^p, \theta_{m}^q\bigr).
\end{aligned}
\label{eq:6}
\end{equation}

This process suppresses the influence of local outliers and improves matching uniformity between the sparse query graph and the dense prior graph.
That is, it alleviates the limitations of the subgraph extraction method proposed in previous work~\cite{goreloc}, which is sensitive to structural discrepancies.
\subsection{Global Localization}
\subsubsection{Candidate Node-pair Extraction}
In large-scale prior maps, the extraction of candidate nodes reduces complexity and mitigates outlier correspondences, enhancing the stability of alignment-based pose estimation.
We extract the top-$\tau$ prior nodes for each query node using the propagated similarity scores from \ac{CALP}, and generate the candidate node-pair set $\mathcal{P}$ as follows:
\begin{equation}
\mathcal{P} = \bigl\{(o_{i}, d_{j}) \mid \operatorname{rank}\bigl(S(o_{i}, d_{j})\bigr) \leq \tau \bigr\},
\label{eq:7}
\end{equation}
where $\operatorname{rank}(\cdot)$ selects the top $\tau$ prior nodes $o_i$ corresponding to $d_j$ based on $S(o_i, d_j)$, thereby forming a reliable subgraph.
\begin{algorithm}[t]
\small
\caption{Candidate Extraction and Pose Estimation}
\label{alg:stochastic_pose}
\begin{algorithmic}[1]
\renewcommand{\algorithmiccomment}[1]{\hfill\#~#1}
\Require Query graph $\mathcal{G}^q$, prior graph $\mathcal{G}^p$, candidate pair number $\tau$, iteration number $N_{\text{iter}}$
\Ensure Estimated camera pose $T$, correspondence set $\mathcal{P}^*$
\\\# \text{M-LLE} and \text{CALP} are described in Sec.~\ref{sec:3-C} and~\ref{sec:3-D}
\State $\mathcal{P} \gets \text{GetCandidates}(\text{CALP}(\text{M-LLE}(\mathcal{G}^p, \mathcal{G}^q)), \tau) \ \text{using Eq.~\eqref{eq:7}}$
\State $(\mathcal{W}, T, \mathcal{P}^*) \gets (0, \mathrm{None}, \mathrm{None})$
\For{$\text{iter}=1$ \textbf{to} $N_{\text{iter}}$}
  \State $\mathcal{P}_{\text{samp}} \gets \text{RandomSample}(\mathcal{P},3)$ 
  \If{\textbf{not} \text{IsValidSample}($\mathcal{P}_{\text{samp}}$)}
    \State \textbf{continue}
  \EndIf
    \State $\mathcal{T}_{\text{tmp}} \gets \text{P3P}(\mathcal{P}_{\text{samp}})$
    \For{$T_{\text{tmp}} \in \mathcal{T}_{\text{tmp}}$}
      \State $\mathcal{W}_{\text{tmp}}, \mathcal{P}^*_{\text{tmp}} \gets \text{CalculateWAS}(T_{\text{tmp}},\mathcal{P}) \ \text{using Eq.~\eqref{eq:9}}$
      \If{$\mathcal{W}_{\text{tmp}}>\mathcal{W}$}
        \State $(\mathcal{W}, T, \mathcal{P}^*) \gets (\mathcal{W}_{\text{tmp}}, T_{\text{tmp}}, \mathcal{P}^*_{\text{tmp}})$
      \EndIf
    \EndFor
\EndFor
\State \Return $T, \mathcal{P}^* $
\end{algorithmic}
\end{algorithm}
\subsubsection{Stochastic Optimal Pose Estimation}
We adopt a stochastic optimal iterative approach inspired by \cite{goreloc}, as outlined in Algorithm \ref{alg:stochastic_pose}.
At each of the $N_{\text{iter}}$ iterations, three unique node-pairs $\mathcal{P}_{\text{samp}}$ are randomly sampled from $\mathcal{P}$.
Both subgraphs induced by this sampled set $\mathcal{P}_{\text{samp}}$ must preserve identical edge connectivity, and previously used sampled sets are never reused.
Given $\mathcal{P}_{\text{samp}}$ that satisfies the required conditions, provisional camera poses $\mathcal{T}_{\text{tmp}}$ are then estimated by solving the \ac{P3P} problem.
Subsequently, for each $(o_i, d_j) \in \mathcal{P}$, the dual quadric $\mathbf{q}^p_i$ of $o_i$ is transformed by each $T_{\text{tmp}} \in \mathcal{T}_{\text{tmp}}$ to generate a projected prior bounding box $B_i$.
Both $B_i$ and the corresponding query bounding box $B_j$ are modeled as Gaussian distributions, denoted as $D_i = G(\mu_i, \Sigma_i)$ and $D_j = G(\mu_j, \Sigma_j)$, respectively.
The degree of alignment between $D_i$ and $D_j$ is measured by the normalized Wasserstein distance proposed by \cite{wang2021normalized} as follows: 
\begin{equation}
\label{eq:8}
W_n\bigl(D_i,D_j\bigr) = \exp\left(-\frac{\sqrt{W_2^{2}\bigl(D_i,D_j\bigr)}}{C}\right),
\end{equation}
where $W_2^{2}(D_i,D_j)$ is the $2^{\text{nd}}$ order Wasserstein distance between two Gaussians, and $C$ is a scale factor.
The best correspondences $\mathcal{P}^{*}=\{(o_{i}, d_{j})^*\}$ are selected from $\mathcal{P}$ by maximizing $W_n(D_i, D_j)$ for each query node. The \ac{WAS} is then calculated as follows:
\begin{equation}
\label{eq:9}
\mathcal{W}=\frac{1}{|\mathcal{P}^{*}|}\sum_{(o_{i}, d_{j})^*\in\mathcal{P}^{*}}
W_n\bigl(D_i,D_j\bigr).
\end{equation}
After all iterations, the algorithm returns the camera pose $T$ and $\mathcal{P}^{*}$ that achieve the maximum $\mathcal{W}$ as the final solution.

\section{Experiments}
\subsection{Setup}
\subsubsection{Datasets}
In this section, we evaluate the proposed method on real-world TUM RGB-D~\cite{tum} and synthetic ICL-LM~\cite{icl} public benchmarks, as well as a custom resort-interior dataset captured in the real world.
From TUM RGB-D, we use the `Fr2\_desk' and `Fr2\_person' sequences, which share similar layouts, with `Fr2\_person' including dynamic factors.
In the ICL-LM dataset, we select the `Walk', `Head', `VR', `DJI', and `Ground' sequences.
These sequences involve complex object layouts with ambiguous classes, captured by diverse platforms in a virtual indoor setting.
The custom resort dataset includes rooms where many objects are partially occluded by walls. 
It is divided into `Resort\_1' and `Resort\_2', which feature stable and dynamic camera movements, respectively.
Object-level maps are built from `Fr2\_desk', `Walk', and `Resort\_1', with data association and pose estimation evaluated on all sequences.
\subsubsection{Method Configurations}
We adopt SH~\cite{semantichistogram} and GOReloc~\cite{goreloc} as baselines for comparison. Both methods rely on single-label, graph-based descriptors.
Experiments were conducted under the following configurations:
\begin{itemize}
\item \textbf{Closed-set:} We utilize a YOLOv8~\cite{yolo} model trained on COCO~\cite{coco} and LVIS~\cite{lvis} datasets to detect and classify objects within a predefined set of categories. This configuration is denoted as Y.
\item \textbf{Open-set:} We employ Grounding DINO~\cite{groundingdino} for zero-shot object detection, and integrate foundation models such as \ac{OVSAM}~\cite{ovsam} and \ac{TAP}~\cite{tap} for zero-shot classification. These configurations are denoted as G+O and G+T, respectively.
\end{itemize}
\subsubsection{Implementation Details}
To construct prior maps, we adopt an object-based SLAM framework~\cite{quadricslam} that leverages 2D bounding-box detections.
We evaluate all methods on COCO (80 classes) and LVIS (1203 classes) categories.
SH is evaluated on COCO only due to computational constraints.

Detection confidence thresholds are set to 0.1 for YOLOv8 and 0.14 for Grounding DINO, with non-maximum suppression applied at an overlap ratio above 0.6.
Hyperparameters are determined empirically based on preliminary experiments: we set $K=5$ in Eq. \eqref{eq:2}, $\tau=3$ in Eq. \eqref{eq:7}, and $C=100$ in Eq. \eqref{eq:8} for all experiments unless otherwise specified.
\subsection{Evaluation Metrics}
\subsubsection{Precision, Recall, and F1-score}
We evaluate multi-object data association by projecting associated objects onto the image plane based on the ground-truth pose. 
Precision, Recall, and F1-score are defined as:
\begin{equation}
P = \frac{\text{TP}}{\text{TP} + \text{FP}}, \quad
R = \frac{\text{TP}}{\text{TP} + \text{FN}}, \quad
F_1 = \frac{2 \cdot P \cdot R}{P + R},
\end{equation}
where TP, FP, and FN denote true positives, false positives, and false negatives, respectively.
\subsubsection{Multiple Object Tracking Accuracy}
To evaluate the temporal consistency of data association across frames, we employ \ac{MOTA}:
\begin{equation}
\text{MOTA} = 1 - \frac{\sum_{t} ( \text{FN}_t + \text{FP}_t + \text{IDS}_t )}{\sum_{t} \text{GT}_t},
\end{equation}
where $\text{FN}_t$, $\text{FP}_t$, $\text{IDS}_t$, and $\text{GT}_t$ denote false negatives, false positives, ID switches, and ground-truth detections at frame $t$. The association cost is defined in Eq. \eqref{eq:8}.
\subsubsection{Translation Error and Success Rate}
For pose estimation, we use \ac{TE}:
\begin{equation}
\text{TE} = \left\lVert \hat{t}-t \right\rVert_2,
\end{equation}
where $\hat{t}$ and $t$ are the estimated and ground-truth positions, respectively.
The mean \ac{TE}, calculated over all success frames, is denoted as $\overline{\text{TE}}$.
The \ac{SR} is the fraction of poses with TE below a threshold (@TE): $\text{SR}_\text{succ}$ for success frames, $\text{SR}_\text{all}$ for all frames.
\subsubsection{Shannon Entropy}
To quantify frame-wise semantic uncertainty, we compute Shannon entropy for each object:
\begin{equation}
H(\hat{\mathcal{C}}_j) = - \sum_{v=1}^{K} \hat{c}_{j,v} \log \hat{c}_{j,v},
\end{equation}
where $\hat{c}_{j,v}$ is the normalized confidence score corresponding to the label $l^q_{j,v}$ in $\hat{\mathcal{C}}_j$, as detailed in Sec.~\ref{sec:3-B}.
\subsection{Data Association}\label{sec:4-B}
\subsubsection{Semantic Disambiguation}
\begin{figure}[t]
	\centering
	\def\width{0.45\textwidth}%
    {%
        \includegraphics[clip, trim= 0 0 0 5, width=\width]{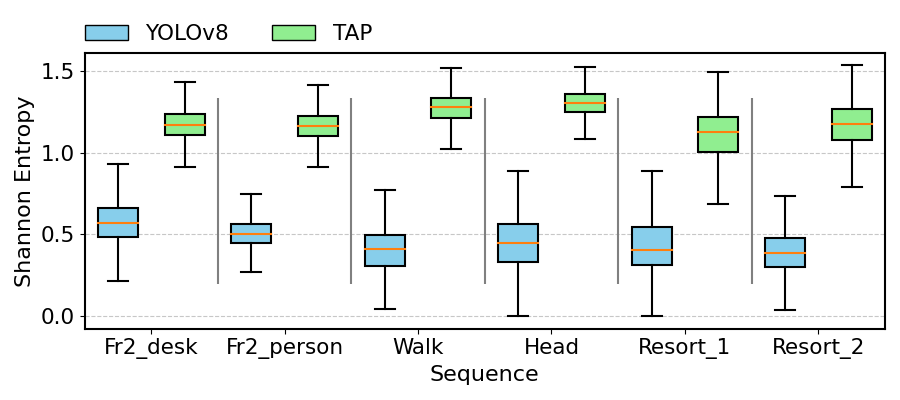}
    }
        \vspace{-0.4cm}
        \caption{\textbf{Frame-wise semantic uncertainty analysis.} 
Shannon entropy is calculated for each object using the normalized confidence scores of its top-5 predicted LVIS~\cite{lvis} labels.
The closed-set detector (YOLOv8~\cite{yolo}) consistently produces low entropy, as it tends to allocate most of its confidence to a single predicted label. Conversely, the open-set detector with a zero-shot classifier (TAP~\cite{tap}) produces a confidence distribution spanning multiple labels, resulting in higher entropy.
}
\label{fig:fig4}
\vspace{-0.5cm}
\end{figure}

As shown in Table~\ref{tab:table1}, MSG-Loc consistently outperforms SH and GOReloc in both $F_1$ and MOTA. 
Averaged over all sequences under both Y and G+T configurations, MSG-Loc achieves relative improvements of 10.6\% in $F_1$ and 33.4\% in MOTA compared to the best baseline.
Interestingly, while MSG-Loc maintains stable performance across all configurations, the baselines tend to degrade under the open-set configuration.
As illustrated in Fig.~\ref{fig:fig4}, the open-set configuration distributes confidence across multiple labels, reflecting broader semantic uncertainty.
Consequently, single-label approaches exhibit increased noise due to label inconsistency.
In contrast, the proposed method suppresses such noise by leveraging both semantic and topological contexts, suggesting that the multi-label likelihood can effectively handle semantic uncertainty regardless of the underlying classification paradigm.
\subsubsection{Category Scalability}
\begin{table}[t]
    \centering
    \caption{Performance Comparison on COCO Categories}
    \resizebox{0.49\textwidth}{!}{
    \begin{tabular}{l l c c c c c c}
        \toprule
        \multirow{2}{*}{Sequence} & \multirow{2}{*}{Method} & \multirow{2}{*}{$F_1\uparrow$} & \multirow{2}{*}{MOTA$\uparrow$} & \multicolumn{3}{c}{SR$_\text{succ}$ [\%]$\uparrow$} & \multirow{2}{*}{$\overline{\text{TE}}$ [m]$\downarrow$} \\
        \cmidrule(r){5-7}
         &  &  &  & @0.5 m & @1 m & @2 m &  \\
        \midrule
        \multirow{7}{*}{Fr2\_desk}        
            & SH (Y)        & 0.905 & 0.786 & 75.08 & 80.30 & 86.13 & 0.674 \\
            & GOReloc (Y)        & 0.882 & 0.739 & 69.34 & 75.73 & 81.65 & 0.845 \\
            & MSG-Loc (Y)  & \textbf{0.942} & \textbf{0.873} & \textbf{88.85} & \textbf{91.96} & \textbf{95.64} & \textbf{0.330} \\
            \cmidrule(lr){2-8}
            & SH (G+T)        & 0.898 & 0.774 & 77.34 & 81.14 & 87.00 & 0.655 \\
            & GOReloc (G+T)        & 0.899 & 0.771 & 79.01 & 83.61 & 88.27 & 0.581 \\
            & MSG-Loc (G+O)        & 0.942 & 0.868 & 88.25 & 91.94 & 96.27 & 0.336 \\            
            & MSG-Loc (G+T)    & \textbf{0.957} & \textbf{0.909} & \textbf{94.37} & \textbf{95.99} & \textbf{97.46} & \textbf{0.231} \\
        \midrule
        \multirow{7}{*}{Walk}
            & SH (Y)        & 0.847 & 0.633 & 55.11 & 61.97 & 65.85 & 3.039\\
            & GOReloc (Y)        & 0.861 & 0.668 & 51.95 & 61.98 & 69.53 & 2.623 \\
            & MSG-Loc (Y)        & \textbf{0.957} & \textbf{0.882} & \textbf{80.16} & \textbf{87.98} & \textbf{90.43} & \textbf{0.922} \\
            \cmidrule(lr){2-8}
            & SH (G+T)        & 0.801 & 0.541 & 40.57 & 51.58 & 58.12 & 4.610 \\
            & GOReloc (G+T)        & 0.831 & 0.595 & 45.64 & 54.05 & 60.48 & 4.254 \\
            & MSG-Loc (G+O)        & 0.938 & 0.829 & 69.92 & 83.14 & 88.91 & 1.240 \\              
            & MSG-Loc (G+T)      & \textbf{0.940} & \textbf{0.860} & \textbf{80.62} & \textbf{93.86} & \textbf{95.85} & \textbf{0.595} \\
        \midrule
        \multirow{7}{*}{Resort\_1}
            & SH (Y)        & 0.909 & 0.786 & 65.32 & 74.19 & 81.45 & 1.001 \\
            & GOReloc (Y)        & 0.932 & 0.833 & 61.65 & 81.20 & 88.72 & 0.835 \\
            & MSG-Loc (Y)        & \textbf{0.959} & \textbf{0.907} & \textbf{74.38} & \textbf{89.16} & \textbf{95.57} & \textbf{0.514} \\
            \cmidrule(lr){2-8}
            & SH (G+T)        & 0.751 & 0.472 & 33.33 & 46.22 & 52.94 & 2.482 \\
            & GOReloc (G+T)        & 0.731 & 0.432 & 32.85 & 44.19 & 52.03 & 2.636 \\
            & MSG-Loc (G+O)        & 0.899 & 0.780 & \textbf{78.85} & \textbf{88.93} & \textbf{93.28} & \textbf{0.578} \\              
            & MSG-Loc (G+T)      & \textbf{0.945} & \textbf{0.880} & 76.83 & 84.75 & 88.91 & 0.789 \\
        \bottomrule
        \addlinespace[2pt]
        \multicolumn{8}{r}{\textbf{Bold}: Best performance.} 
    \end{tabular}%
    }
    \label{tab:table1}
\end{table}
Under COCO-to-LVIS scaling, Table~\ref{tab:table2} indicates that MSG-Loc improves MOTA by 9.3\% on average across all sequences when transitioning from closed-set (Y) to open-set (G+T). 
In contrast, GOReloc shows reduced association performance in most sequences under the same transition.
As category diversity increases, single-label approaches tend to be more susceptible to misclassifications. 
The proposed multi-label approach preserves a broader set of semantic candidates, reducing misclassification and improving inter-object discriminability. 
Additionally, MSG-Loc (G+O) generally achieves lower performance than MSG-Loc (G+T) across both COCO and LVIS categories, suggesting that multi-label predictions generated by \ac{OVSAM} may be less stable than those produced by \ac{TAP}.

Fig.~\ref{fig:fig5}(a) qualitatively illustrates the data association performance of the proposed method.
In cluttered indoor scenarios, MSG-Loc (G+T) successfully matches multiple ambiguous objects, whereas the baseline occasionally confuses similar ones, resulting in matching errors.
\subsection{Pose Estimation}
\begin{table}[t]
    \centering
    \caption{Performance Comparison on LVIS Categories}
    \resizebox{0.49\textwidth}{!}{
    \begin{tabular}{l l c c c c c c}
        \toprule
        \multirow{2}{*}{Sequence} & \multirow{2}{*}{Method} & \multirow{2}{*}{$F_1\uparrow$} & \multirow{2}{*}{MOTA$\uparrow$} & \multicolumn{3}{c}{SR$_\text{succ}$ [\%]$\uparrow$} & \multirow{2}{*}{$\overline{\text{TE}}$ [m]$\downarrow$} \\
        \cmidrule(r){5-7}
         &  &  &  & @0.5 m & @1 m & @2 m &  \\
        \midrule
        \multirow{5}{*}{Fr2\_desk}        
            & GOReloc (Y)        & 0.951 & 0.895 & 86.36 & 89.99 & 93.69 & 0.429 \\
            & MSG-Loc (Y)        & \textbf{0.970} & \textbf{0.936} & \textbf{96.23} & \textbf{97.40} & \textbf{98.19} & \textbf{0.211} \\
            \cmidrule(lr){2-8}
            & GOReloc (G+T)        & 0.906 & 0.790 & 74.13 & 80.53 & 87.49 & 0.673 \\
            &MSG-Loc (G+O)        & 0.918 & 0.822 & 91.34 & 96.56 & 98.84 & 0.245 \\            
            &MSG-Loc (G+T)      & \textbf{0.976} & \textbf{0.947} & \textbf{97.40} & \textbf{98.28} & \textbf{99.58} & \textbf{0.159} \\
        \midrule
        \multirow{5}{*}{Fr2\_person}        
            & GOReloc (Y)        & \textbf{0.894} & \textbf{0.776} & 57.13 & 73.90 & \textbf{81.71} & \textbf{1.008} \\
            & MSG-Loc (Y)        & 0.875 & 0.730 & \textbf{58.84} & \textbf{74.27} & 78.56 & 1.044 \\
            \cmidrule(lr){2-8}
            & GOReloc (G+T)        & 0.754 & 0.448 & 38.73 & 45.06 & 54.12 & 2.304 \\
            & MSG-Loc (G+O)        & 0.850 & 0.631 & 62.16 & 68.73 & 73.81 & 1.079 \\            
            & MSG-Loc (G+T)      & \textbf{0.901} & \textbf{0.773} & \textbf{76.07} & \textbf{79.00} & \textbf{82.54} & \textbf{0.777} \\
        \midrule
        \multirow{5}{*}{Walk}
            & GOReloc (Y)        & 0.886 & 0.751 & 63.47 & 77.70 & 86.78 & 1.283 \\
            & MSG-Loc (Y)        & \textbf{0.926} & \textbf{0.838} & \textbf{78.10} & \textbf{90.84} & \textbf{95.33} & \textbf{0.576} \\
            \cmidrule(lr){2-8}
            & GOReloc (G+T)        & 0.914 & 0.809 & 64.51 & 74.35 & 80.07 & 2.385 \\
            & MSG-Loc (G+O)        & 0.951 & 0.885 & 72.34 & 86.04 & 94.82 & 0.716 \\              
            & MSG-Loc (G+T)      & \textbf{0.961} & \textbf{0.906} & \textbf{84.82} & \textbf{94.24} & \textbf{98.25} & \textbf{0.426} \\
        \midrule
        \multirow{5}{*}{Head}
            & GOReloc (Y)        & 0.789 & 0.575 & 29.01 & 44.17 & 58.02 & 3.726 \\
            & MSG-Loc (Y)        & \textbf{0.895} & \textbf{0.774} & \textbf{42.58}& \textbf{59.32} & \textbf{73.13} & \textbf{2.403} \\
            \cmidrule(lr){2-8}
            & GOReloc (G+T)        & 0.775 & 0.548 & 20.84 & 33.63 & 43.96 & 6.537 \\
            & MSG-Loc (G+O)        & 0.872 & 0.702 & 27.98 & 46.49 & 63.10 & 3.895 \\              
            &MSG-Loc (G+T)      & \textbf{0.913} & \textbf{0.791} & \textbf{44.64} & \textbf{63.63} & \textbf{76.57} & \textbf{2.493} \\
        \midrule
        \multirow{5}{*}{Resort\_1}
            & GOReloc (Y)        & 0.828 & 0.624 & 61.83 & 77.75 & 89.23 & 0.879 \\
            & MSG-Loc (Y)        & \textbf{0.883} & \textbf{0.759} & \textbf{80.43} & \textbf{93.99} & \textbf{96.32} & \textbf{0.432} \\
            \cmidrule(lr){2-8}
            & GOReloc (G+T)        &0.893  &0.744  &57.58  &69.70  &78.45  &1.190  \\
            & MSG-Loc (G+O)        & 0.952 & 0.895 & 81.58 & 90.50 & 95.45 & 0.488 \\              
            & MSG-Loc (G+T)      & \textbf{0.953} & \textbf{0.900} & \textbf{88.01} & \textbf{94.51} & \textbf{95.73} & \textbf{0.398} \\
        \midrule
        \multirow{5}{*}{Resort\_2}
            & GOReloc (Y)        & \textbf{0.820} & \textbf{0.620} & 58.27 & \textbf{81.77} & \textbf{90.04} & \textbf{0.853} \\
            & MSG-Loc (Y)        & 0.815 & 0.604 & \textbf{62.03} & 77.82 & 88.91 & 0.882 \\
            \cmidrule(lr){2-8}
            & GOReloc (G+T)        &0.634  &0.262  &19.93  &34.78  &42.03  &3.273  \\
            & MSG-Loc (G+O)       & 0.832 & 0.648 & \textbf{54.46} & \textbf{65.10} & \textbf{73.76} & \textbf{1.526} \\              
            & MSG-Loc (G+T)      & \textbf{0.867} & \textbf{0.724} & 54.28 & 61.95 & 71.98 & 1.614 \\
        \bottomrule
    \end{tabular}%
    }
    \label{tab:table2}
\end{table}
\subsubsection{Pose Accuracy}
As shown in Tables~\ref{tab:table1} and~\ref{tab:table2}, MSG-Loc demonstrates higher SR$_\text{succ}$ and lower $\overline{\text{TE}}$ than the baselines in most cases, including complex scenes.
In the `Fr2\_desk', `Walk', and `Resort\_1' sequences, our method achieves a success rate above 70\% at the 0.5 m threshold and $\overline{\text{TE}}$ below 1 m across both categories.
The baselines construct their descriptors from all edge-connected neighbors using a single-label representation, which can introduce noise from misclassifications during matching.
In contrast, the proposed method selectively integrates only the most probable neighboring node-pairs, indirectly guiding the selection of objects with minimal geometric error relative to the observations.
\begin{figure*}[t]
	\centering
	\def\width{0.932\textwidth}%
    {%
        \includegraphics[clip, trim= 8 10 3 8, width=\width]{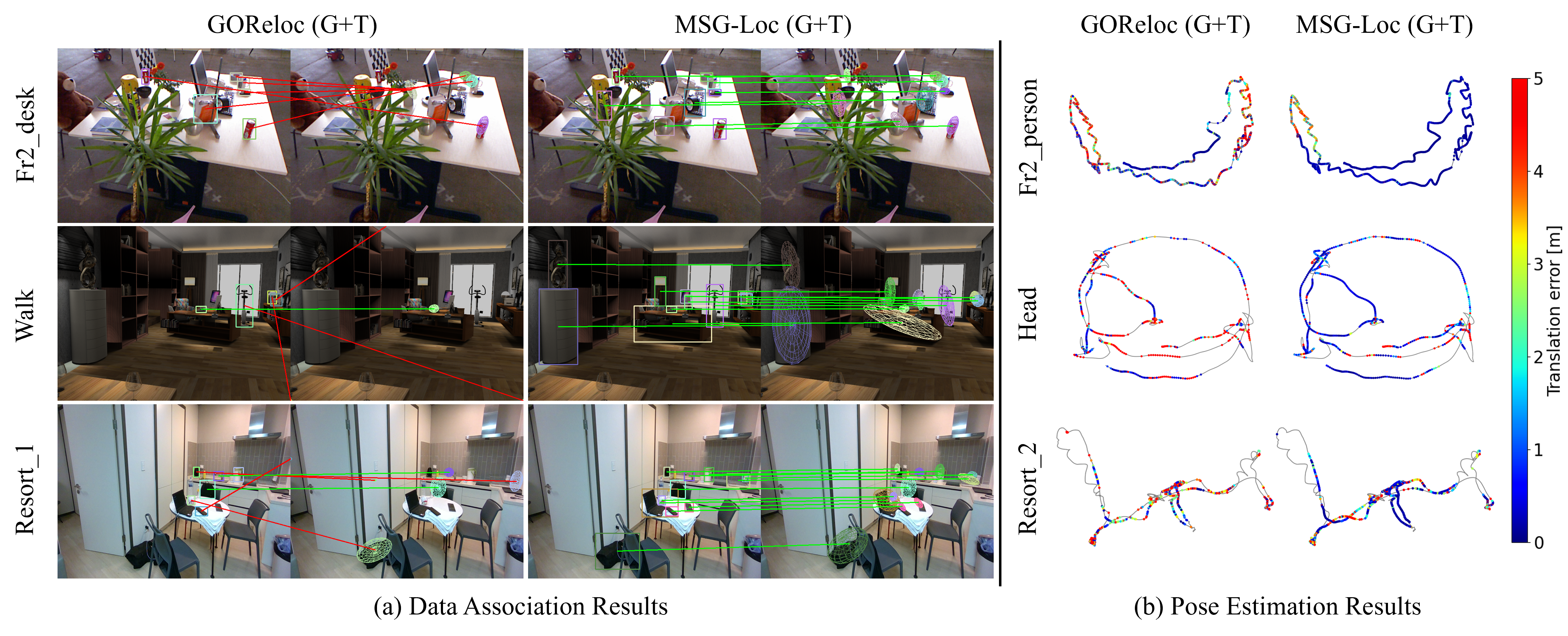}
    }
    \vspace{-0.2cm}
    \caption{\textbf{The qualitative results of data association and pose estimation.}
(a) For each pair, the query frame image is presented on the left, and the corresponding landmark rendering is on the right. Green lines represent correct associations, whereas red lines indicate incorrect ones.
(b) The color bar on the right denotes the magnitude of translation error. 
Scatter points represent successful pose estimates and are colored according to their translation error as indicated by the color bar, whereas continuous gray lines denote failures.
} 
    \label{fig:fig5}
    \vspace{-0.5cm}
\end{figure*}
\subsubsection{Open-set Stability of Localization}
GOReloc shows a large performance gap between the closed-set and open-set configurations across most sequences.
These fluctuations stem from the graph kernel-based approach, which amplifies mismatches under label noise. 
This effect becomes more evident when partial observations are frequent and object detections are sparse.
In contrast, MSG-Loc maintains comparatively stable pose estimation performance across diverse sequences.
Notably, MSG-Loc (G+T) achieves additional improvements in all metrics when scaling from COCO to LVIS.

To qualitatively demonstrate these advantages, Fig.~\ref{fig:fig5}(b) visualizes pose estimation results across three complete sequences. 
Gray lines denote frames that failed due to sparse detections or invalid correspondence graphs, while dense scatter points indicate a high rate of successful and precise localization.
This highlights the robustness of MSG-Loc under various conditions, including category expansion.
\begin{table}[t]
    \centering
    \caption{Comparison of Pose Estimation with Feature-based Method}
    \resizebox{0.40\textwidth}{!}{%
    \begin{tabular}{l l c c c  c}
        \toprule
        \multirow{2}{*}{Sequence} & \multirow{2}{*}{Method} & \multicolumn{4}{c}{\makecell{SR$_\text{all}$ [\%]$\uparrow$}}\\
        \cmidrule(r){3-6} 
        &  & @0.5 m & @1 m & @2 m &@5 m\\
        \midrule
        \multirow{4}{*}{Fr2\_person} 
            & ORB-SLAM2            &8.68  &8.68  &8.68  &8.68 \\
            & SH w/ O             &36.59  &41.87  &48.66  &75.02  \\
            & GOReloc w/ O             &41.78  &48.46  &57.14  &80.30  \\
            & MSG-Loc w/ O    &\textbf{59.48} & \textbf{61.54} & \textbf{66.56}  &\textbf{87.44}\\
    
        \midrule
        \multirow{4}{*}{Head} 
            & ORB-SLAM2            &47.82  &48.97  &48.97  &48.97 \\
            & SH w/ O             &49.05  &50.65  &53.40  &58.59  \\        
            & GOReloc w/ O              &49.43  &51.72  &55.23   &61.12 \\
            & MSG-Loc w/ O    &\textbf{57.45} & \textbf{63.33} & \textbf{67.91}  &\textbf{72.04}\\
        \midrule
        \multirow{4}{*}{VR} 
            & ORB-SLAM2            &26.98  &27.12  &27.12  &27.12 \\
            & SH w/ O             &27.46  &28.56  &30.87  &39.56  \\ 
            & GOReloc w/ O              &27.75  &29.43  &32.26   &41.48 \\
            & MSG-Loc w/ O    &\textbf{30.10}  &\textbf{34.04}  &\textbf{37.30}  &\textbf{44.46} \\
            \midrule
        \multirow{4}{*}{DJI} 
            & ORB-SLAM2            &5.90  &5.90  &5.90  &5.90  \\
            & SH w/ O             &6.80  &9.65  &16.79  &36.98  \\          
            & GOReloc w/ O              &8.20  &11.24  &16.44  &32.08 \\
            & MSG-Loc w/ O   &\textbf{12.54} &\textbf{17.19} &\textbf{22.89} &\textbf{33.73}\\
        \midrule
        \multirow{4}{*}{Ground} 
            & ORB-SLAM2            &2.56  &2.75  &2.75  &2.75  \\
            & SH w/ O             &\textbf{3.69}  &\textbf{4.31}  &5.06  &\textbf{10.87}  \\           
            & GOReloc w/ O              &2.94  &3.62  &4.88  &10.56 \\
            & MSG-Loc w/ O    &2.62 & 3.50 & \textbf{5.75} &\textbf{10.87}\\
        \midrule
        \multirow{4}{*}{Resort\_2} 
            & ORB-SLAM2            &8.48  &15.18  &16.22   &16.22 \\
            & SH w/ O             &9.09  &16.34  &18.41  &23.66  \\          
            & GOReloc w/ O              &8.96  &16.04  &18.23   &23.78 \\ 
            & MSG-Loc w/ O    &\textbf{12.99} &\textbf{20.30}  &\textbf{22.93}   &\textbf{29.15} \\
        \bottomrule
        \addlinespace[2pt]
        \multicolumn{6}{r}{
        ‘w/ O’ denotes feature augmentation with ORB-SLAM2.
        } 
    \end{tabular}%
    }
    \label{tab:table3}
\end{table}
\subsubsection{Scene Invariance}
Table~\ref{tab:table3} presents a comparison of ORB-SLAM2 with the feature-augmented variants of SH, GOReloc, and MSG-Loc under the open-set configuration with COCO categories.
The integrated approaches consistently outperform the purely feature-based ORB-SLAM2 across all evaluated sequences.
On the more ambiguous ICL-LM sequences, the baselines tend to remain comparable to ORB-SLAM2 at the strict 0.5 m threshold, in contrast to MSG-Loc.
Notably, our method achieves successful pose estimation within 0.5 m TE for more than 50\% of frames on the `Fr2\_person' and `Head' sequences.
These results indicate that multi-label reasoning improves inlier discriminability across varying scene conditions and viewpoints, thereby increasing the overall pose-estimation success rate over all frames.
\begin{table}[t]
    \centering
    \caption{Ablation Study on Top-$\tau$ under Open-set Configuration}
    \resizebox{0.47\textwidth}{!}{
    \begin{tabular}{c l c c c c c c}
        \toprule
        \multirow{2}{*}{Top-$\tau$} & \multirow{2}{*}{Method} &
        \multicolumn{3}{c}{Fr2\_person: SR$_\text{succ}$ [\%]$\uparrow$} &
        \multicolumn{3}{c}{Head: SR$_\text{succ}$ [\%]$\uparrow$} \\
        \cmidrule(r){3-5}\cmidrule(l){6-8}
        & & @0.5 m & @1 m & @2 m & @0.5 m & @1 m & @2 m \\
        \midrule
        \multirow{4}{*}{$\tau=1$} 
        & SH   & 34.07 & 41.50 & 50.98 & 6.46 & 11.73 & 18.03 \\
        & GOReloc   & 31.71 & 38.78 & 51.02 & 6.43 & 11.04 & 19.77 \\
        & M-LLE    & 49.87 & 56.33 & \textbf{66.66} & 32.04 & 42.96 & 54.79 \\
        & M-LLE + CALP    & \textbf{53.45} & \textbf{58.00} & 64.65 & \textbf{35.97} & \textbf{49.11} & \textbf{60.80} \\
        \midrule
          \multirow{4}{*}{$\tau=3$} 
        & SH   & 40.99 & 48.04 & 56.85 & 7.44 & 12.36 & 19.79 \\
        & GOReloc   & 47.12 & 55.58 & 66.12 & 8.32 & 14.65 & 23.42 \\
        & M-LLE    & 62.51 & 68.72 & 74.67 & 30.82 & 43.50 & 53.77 \\
        & M-LLE + CALP    & \textbf{67.05} & \textbf{69.38} & \textbf{75.07} & \textbf{36.07} & \textbf{49.36} & \textbf{60.02} \\
        \bottomrule
    \end{tabular}}
    \label{tab:table4}
\end{table}
\begin{table}[t]
    \centering
    \caption{Ablation Study on Top-$K$ under Open-set Configuration}
    \resizebox{0.47\textwidth}{!}{
    \begin{tabular}{l c c c c c c c}
        \toprule
        \multirow{2}{*}{Method} & \multirow{2}{*}{Top-$K$} &
        \multicolumn{3}{c}{Fr2\_person: SR$_\text{succ}$ [\%]$\uparrow$} &
        \multicolumn{3}{c}{Head: SR$_\text{succ}$ [\%]$\uparrow$} \\
        \cmidrule(r){3-5}\cmidrule(l){6-8}
        & & @0.5 m & @1 m & @2 m & @0.5 m & @1 m & @2 m \\
        \midrule
        \multirow{3}{*}{M-LLE} 
        & $K=1$    & 30.30 & 36.66 & 47.41 & 13.33 & 21.05 & 25.88 \\
        & $K=3$   & 72.23 & 77.30 & \textbf{81.73} & 36.90 & 52.86 & 64.40 \\
        & $K=5$   & \textbf{73.95} & \textbf{77.59} & 81.43 & \textbf{38.45} & \textbf{59.88} & \textbf{74.32} \\
        \midrule
          \multirow{3}{*}{M-LLE + CALP} 
        & $K=1$   & 54.03 & 59.61 & 65.38 & 32.78 & 44.65 & 51.56 \\
        & $K=3$   & 75.60 & 78.85 & 82.40 & 43.91 & 60.22 & 70.90 \\
        & $K=5$   & \textbf{76.07} & \textbf{79.00} & \textbf{82.54} & \textbf{44.64} & \textbf{63.63} & \textbf{76.57} \\
        \bottomrule
    \end{tabular}}
    \label{tab:table5}
\end{table}
\subsection{Ablation Study}
\subsubsection{Analysis of the Correspondence Candidate Set}
Table~\ref{tab:table4} presents an ablation study on Top-$\tau$ under the COCO open-set configuration, comparing the baselines with the proposed modules.
As $\tau$ increases from 1 to 3, SR$_\text{succ}$ generally improves across most methods. While M-LLE shows slight regressions in a few cases, its combination with CALP yields consistent improvements across both sequences and most thresholds. 
These results indicate that expanding the candidate set necessitates topology-based consistency constraints to effectively suppress noisy candidates and ensure valid inference.
\subsubsection{Analysis of Semantic Context}
Table~\ref{tab:table5} reports a Top-$K$ ablation of the proposed modules under the LVIS open-set configuration.
Compared to $K=1$, M-LLE achieves a substantial performance gain at $K=3$, improving SR$_\text{succ}$ by an average of 35.1 percentage points across both sequences and thresholds.
This suggests that retaining more semantic candidates enables better exploitation of semantic context.
Under the same setting, M-LLE + CALP demonstrates consistent improvements as $K$ increases from 1 to 5 by leveraging per-object multi-label predictions and graph-based context to enhance candidate reliability and suppress noise.
Consequently, integrating \ac{M-LLE} and \ac{CALP} improves the stability and generalization of global localization.
\subsection{Runtime Analysis}
All experiments were conducted on a desktop with an Intel i9-13900K CPU.
The runtime comparison is summarized in Table~\ref{tab:table6}.
Despite maintaining high semantic discriminability, MSG-Loc achieves real-time performance, operating above 36.4 Hz across datasets.
GOReloc attains the fastest runtime of about 8.8 ms through a lightweight descriptor-matching design.
In contrast, SH shows notable computational overhead, running below 5.3 Hz due to the heavy cost of histogram-based matching.
This comparison indicates that the proposed method achieves an effective balance between runtime efficiency and semantic correspondence accuracy for online operation.
\begin{table}[tb]
    \centering
    \caption{Runtime Comparison with Baselines}
    \vspace{-0.5em}
    \setlength{\arrayrulewidth}{0.3mm}
    \setlength{\tabcolsep}{4pt}
    \renewcommand{\arraystretch}{1.5}
    \resizebox{0.49\textwidth}{!}{
    \begin{tabular}{l c c c c c c}
        \toprule
         & \multicolumn{3}{c}{TUM RGB-D} & \multicolumn{3}{c}{ICL-LM} \\ 
        \midrule
         & MSG-Loc & GOReloc & SH & MSG-Loc & GOReloc & SH \\
        \midrule
        Preprocessing & 4.17 ms & 4.23 ms & 19.58 ms & 4.38 ms & 4.49 ms & 27.14 ms \\
        Graph Matching & 8.28 ms & 0.50 ms & 168.5 ms & 4.85 ms & 1.10 ms & 454.5 ms \\
        Pose Estimation & 14.95 ms & 4.11 ms & 3.73 ms & 7.30 ms & 3.24 ms & 3.27 ms \\
        \midrule 
        Total
        & 27.40 ms & 8.84 ms & 191.8 ms & 16.53 ms & 8.83 ms & 484.9 ms \\ 

        \bottomrule
    \end{tabular}}
    \vspace{-0.5em}
    \label{tab:table6}
\end{table}
\section{Conclusion}
In this work, we introduce \textit{MSG-Loc}, a novel semantic graph matching framework that leverages multi-label likelihood for object-level global localization.
The proposed method effectively bridges closed-set and open-set paradigms for object detection and classification.
Specifically, M-LLE constructs likelihoods based on the multi-label hypotheses assigned to each observed object and each map landmark, resolving semantic ambiguity without relying on deterministic single-label predictions.
Furthermore, CALP aggregates local maximum likelihoods via the graph structure, enabling consistent data association even with sparse or ambiguous observations.
Experimental results demonstrate the scalability of MSG-Loc, maintaining robust performance even as the number of recognizable categories significantly increases.

In future work, we aim to extend this framework toward an open-world paradigm by incorporating vision-language embeddings and captions for interpretable object localization.

%


\scriptsize
\bibliographystyle{packages/IEEEtranN} 
\bibliography{packages/string-short, packages/references}

\end{document}